\documentclass[10pt,twocolumn,letterpaper]{article}

\usepackage{cvpr}
\usepackage{times}
\usepackage{epsfig}
\usepackage{graphicx}
\usepackage{amsmath}
\usepackage{multirow}
\usepackage{amssymb}
\usepackage{dsfont}
\DeclareMathOperator*{\argmax}{arg\,max}

\usepackage{stmaryrd}
\usepackage{float}
\usepackage{balance}

\usepackage[pagebackref=true,breaklinks=true,letterpaper=true,colorlinks,bookmarks=false]{hyperref}

 \cvprfinalcopy 

\def\cvprPaperID{5074} 

\ifcvprfinal\fi
\begin{document}

\title{A Sample Selection Approach for Universal Domain Adaptation}

\author{Omri Lifshitz\\
School of Computer Science\\
Tel Aviv University\\
\and
Lior Wolf\\
Facebook AI Research and \\
Tel Aviv University\\
}

\maketitle

\begin{abstract}
We study the problem of unsupervised domain adaption in the universal scenario, in which only some of the classes are shared between the source and target domains. We present a scoring scheme that is effective in identifying the samples of the shared classes. The score is used to select which samples in the target domain to pseudo-label during training. Another loss term encourages diversity of labels within each batch. Taken together, our method is shown to outperform, by a sizable margin, the current state of the art on the literature benchmarks. Our code is attached as supplementary and will be released as open source code.
\end{abstract}

\section{Introduction}

In real world situations, the necessity of applying domain adaptation is the rule and not the exception, since ``no man ever steps in the same river twice''. This is true not only for the input samples, whose distribution is likely to change both because of the shifting setting and due to the practical considerations of collecting training samples, but also with regards to the output labels. In many cases, the classes seen and labeled during training differ from those encountered during the deployment phase. 

Unsupervised domain adaptation  seeks to learn a classifier in a source domain in which supervised training samples exist, such that  it would be effective in a target domain for which only unsupervised samples exist. Universal domain adaptation (UDA) adds the challenge that some of the classes in the source domain do not appear in the target domain and vice versa. Therefore, the classifier, when applied to the target domain, has to classify only to the relevant classes, and also identify the samples that belong to the classes that are unique to the target domain.

The method we propose is based on three losses. The first loss is the conventional domain confusion loss, which encourages the representation of the samples to be domain agnostic. The second one is the pseudo-labeling loss, which is a very common loss in semi-supervised learning and, in particular, in unsupervised domain adaptation. However, the application of pseudo-labels in the UDA setting requires additional care, since labeling every sample is almost guaranteed to lead to adverse results. We, therefore, propose to identify the samples in the target domain for which the labels are likely to be in the set of shared classes.

The third loss is the batch diversity loss, which encourages the predicted samples both from the source domain and from the target domain to be uniformly distributed between the classes, and not all predicted to be from the same limited set of classes. This is especially important in the UDA case, in which the learned model can declare all target samples to be from an unknown class. This trivial decision can be justified by the reasoning that a cat in the source domain is not the same as a cat in the target domain. Naturally, diversity among the labels is more justified for correctly classified samples, and we, therefore, similarly to the pseudo-labeling scheme above, attempt to apply it only for samples from the shared classes. 

In our work, the samples from the target domain that are likely to be from the shared classes are identified based on two signals. The first is the certainty of the classifier, assuming that the classifier is more likely to be confused when encountering samples from unseen classes. Second, the samples in the target domain that are more similar to the source domain are likely to be from the shared classes. We, therefore, suggest a scoring scheme that combines the outputs of both the label classifier and the domain classifier.

Our experiments show that using a scoring scheme based on the two aforementioned signals together with the three loss terms improves the state of the art accuracy in the UDA scenario.

Our main contributions are: (i) a direct method for UDA, which employs selective pseudo-labels as the main loss, (ii)  encouraging diversity in the labels by the batch diversity loss, (iii) a new sample scoring scheme that outperforms the one previously proposed, and (iv)
state of the art results across datasets and benchmarks.

\section{Related work}
The problem of unsupervised domain adaptation can be divided into four different categories, based on the relation between the label sets of the source and target domains: closed-set, open-set, partial and universal. \textbf{Closed-set} domain adaptation is a scenario where the source and target domains share the same label set. The main challenge in this scenario is to overcome the \textit{domain gap} that comes as a result of the samples being taken from different distributions. There are two common approaches to the close-set problem: feature adaptation and generative models. Generative based approaches \cite{bousmalis2016unsupervised, sankaranarayanan2017generate, duplex_generative, liu2017detach,murez2017image,swHuang2018, volpi2017adversarial} attempt to generate labeled target samples from the source samples. Methods based on CycleGAN~\cite{zhu2017unpaired} generate synthetic target-like samples from the source domain and source-like samples from the target in order to train classifiers on each of the domains~\cite{hoffman2017cycada,russo2017source}.

Methods based on feature adaptation aim to reduce the discrepancy between the feature distribution of samples from the source and target domains. In \cite{ganin2016domain}, a domain adversarial network is introduced and added to a classifier network with the purpose of creating features that are indiscriminate with respect to a shift between domains, yet still discriminative for the main classification task. By introducing a gradient reversal unit, the feature extractor is trained to produce features that confuse the domain classifier.

\textbf{Open-set} domain adaptation, first proposed by \cite{busto2019open} assumes knowledge of the shared label set between the source and target domain, while all private label sets are marked as ``unknown''. A modification proposed by \cite{saito2018open} requires no data from the private source label set.

\textbf{Partial-set} domain adaptation assumes that the target domain's label set is a subset of the source's label set. Cao \etal~\cite{cao2017partial} employ adversarial distribution matching by using a number of domain discriminators together with a weighting scheme at both the class and instance level. Zhang \etal~\cite{zhang2018importance} use an adversarial  method to identify the source samples that are potentially from the private target label set. The results were further improved by using a single adversarial domain network and down-weighting the data of the source private set for the classifier and domain adversarial during training~\cite{cao2018partial}. 

\textbf{Universal} domain adaptation was first introduced in \cite{UDA_2019_CVPR} and unlike the aforementioned scenarios, it does not assume any prior knowledge about the relation between the source and target label sets.

{\bf Pseudo-labels} are a simple yet effective tool used in closed-set domain adaptation, in order to learn categorical representation of the target domain \cite{french2017selfensembling, saito2017asymmetric, NIPS2016_6360, shu2018dirtt,zhang2018collaborative, choi2019pseudolabeling}. Although the use of pseudo-labels during training can greatly improve the final outcome of the network, the use of false pseudo-labels leads to negative transfer, which is a major concern in UDA.

\section{Problem setting}
\begin{figure*}
  \centering
  \includegraphics[height=0.04\textwidth]{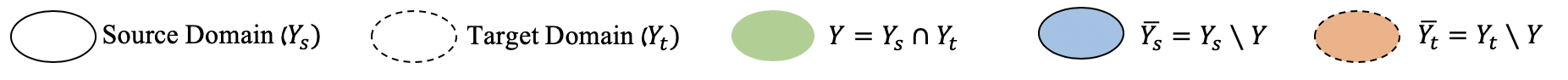}
    \begin{tabular}{ccc}
  \includegraphics[height=0.1\textwidth]{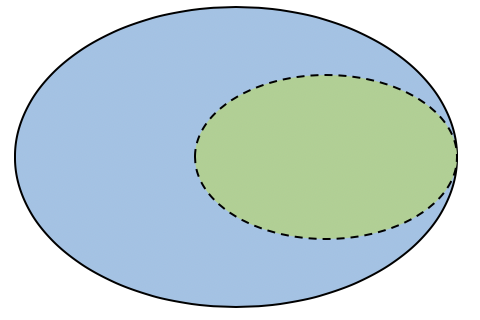} & \includegraphics[height=0.1\textwidth]{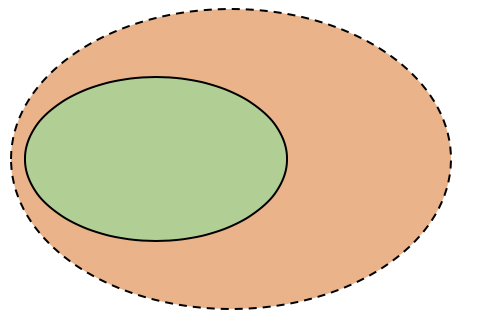} &
  \includegraphics[height=0.1\textwidth]{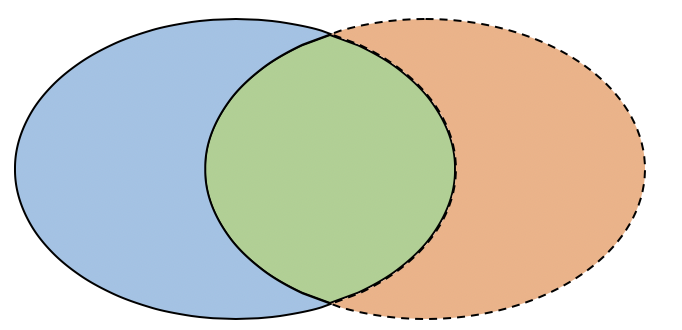} \\
  (a) & (b) & (c)
  \end{tabular}
  \smallskip\caption{Possible label set relations. The UDA framework covers all possible relations between the source and target domain. (a) The case of ``partial domain'', the target label set is a subset of the source label set and thus all target labels are in the shared set. (b) The ``open set'' case, in which the source label set is a subset of the target's. (c) In the general UDA case, the target label set and the source label set intersect, yet both domains have a private label set. In all cases, the shared label set is unknown during training.}
\label{fig:problem_setting}
\end{figure*}

We follow the setting of UDA proposed by \cite{UDA_2019_CVPR}. During training, we are provided with a source domain $D_s = \left\{(x_i^s, y_i^s)\sim p\right\}_{i=1}^{n_s}$ of labeled data sampled from distribution $p$ and a target domain $D_t = \left\{(x_i^t)|\; x_i^t \sim q^x\right\}_{i=1}^{n_t}$ of unlabeled data sampled from distribution $q^x$, which is the marginalization of the distribution $q$ of samples and their labels in the target domain. We denote by $Y_s$ ($Y_t$) the label set of the source (target) domain.  The shared label set is denoted by $Y = Y_s \cap Y_t$. For convenience, we denote the private label sets of the source and target domain in the following manner: $\overline{Y_s} = Y_s \setminus Y$ and $\overline{Y_t} = Y_t \setminus Y$, respectively.

As can be seen in Fig.~\ref{fig:problem_setting}, UDA generalizes all other variants of domain adaptation. Namely, the partial-set case in which the target classes are a subset of the source classes (closed-set is a special case of partial-set), and the open-set case in which the source classes are a subset of the target classes. The latter case is the more challenging of the two since some of the target domain samples cannot be adapted to match the samples seen during training.

The Jaccard index of the label sets of the two domains, $\xi = \frac{Y}{Y_s \cup Y_t}$, is used to measure the overlap in classes. The objective in the UDA scenario is to create a model $g$ that maximises the target classification on the shared label set, as well as distinguishes between samples with labels from $Y$ and those in $\overline{Y_t}$. i.e. 
\begin{equation}
\max_g E_{(x,y)\sim q}[g(x) = t(y)],
\end{equation}
where
\begin{equation}
t(y) = \begin{cases} y &\mbox{if } y\in Y \\
\tau & \mbox{if } y\in \overline{Y_t} \end{cases}
\end{equation}
and $\tau$ is the symbol used to mark unknown classes not seen in the labeled training set $D_s$.

\section{Method}

\begin{figure*}
  \centering
    \begin{tabular}{cc}
  \includegraphics[height=0.152\textwidth]{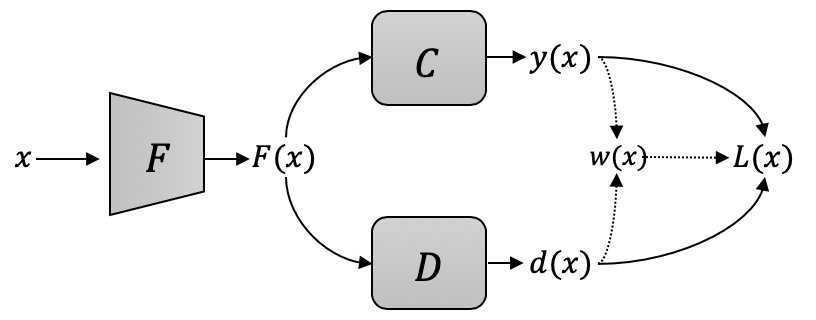} & \includegraphics[height=0.152\textwidth]{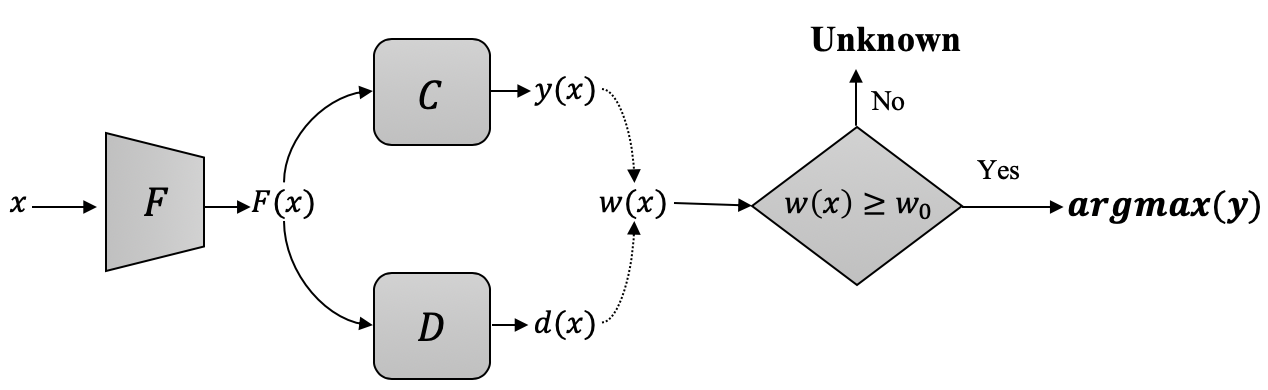} \\
  (a) & (b)
  \end{tabular}
  \smallskip\caption{Architecture of the network during training and deployment. During the training stage (a), the score along with the label and domain classification is used to calculate the loss. During the deployment stage (b) the scores are used as a threshold to decide whether the sample is from the shared label set or should be marked as unknown.}
\label{fig:arch}
\end{figure*}

The architecture we employ is shown in Fig.~\ref{fig:arch}(a). It consists of a domain classifier $D$, a feature extractor $F$, and a label classifier $C$. By using one adversarial domain classifier $D$, our method is simpler than previous work~\cite{UDA_2019_CVPR}, which uses two domain classifiers.

Input $x$ (from both domains) is fed into the feature extractor $F$, yielding the feature vector $F(x)$. $F(x)$ is, in turn, fed to both the domain classifier $D$ and the label classifier $C$. The label classifier outputs the label prediction of classes from the source domain $\bar{y}(x) = C(F(x)) \in \mathbb{R}^{|Y_s|}$, which is a vector of pseudo probabilities obtained by the softmax function. The adversarial domain classifier yields the probability of the sample being from the source domain $d(x) = D(F(x)) \in [0,1]$. The results from both classifiers are used for calculating the \textit{sample transfer score} and for calculating the losses. 

The sample transfer score, $w(x)$, estimates the confidence that $x$ is from the shared label set. The score is calculated using the prediction $\max\bar y(x)$ and the domain classification $d(x)$ as detailed below. A higher value of $w(x)$ indicates that the $x$ appears to be from the shared label set and that the correct label was identified. 

During the deployment stage, the test sample undergoes the same path as before, but rather than calculating losses, we use the score $w(x)$ as a threshold to decide whether we should predict a class or label the sample as the symbol $\tau$ that represents all labels unseen during training. We use a hyper-parameter $w_0$ and output the class label according to the following:
\begin{equation}
    y(x) = \begin{cases} 
        \argmax{\bar{y}} & w(x) > w_0 \\
        \tau & \mbox{otherwise}  
        \end{cases}
        \label{eq:taucase}
\end{equation}

\subsection{The sample transfer score}

We define a scoring mechanism that represents the confidence that a sample $x$ is from the shared label set $Y$. This score is used in both training and deployment. During the training stage, the scores are used as a threshold for losses on samples from the target domain, as explained in the following sections. During the deployment stage, the scores are used in order to decide whether or not a sample should be labeled as $\tau$ or predicted from one of the classes in the source label set, as shown in Eq.~\ref{eq:taucase}.

The score is a combination of two signals: (i) the confidence in the classification label, as it manifests itself in the vector of pseudo probabilities $\bar y(x)$, and (ii) the estimation of the probability of it being in the source domain, as is estimated by $d(x)$. The usage of the second signal on target domain samples, is meant to measure the similarity of these samples to the source domain samples. Naturally, target samples that are more similar to the source domain samples are more likely to be in the shared label set.

It is reasonable to expect that
\begin{equation}
\begin{split}
\mathbb{E}_{(x,y) \in p} \max \bar y(x) > \mathbb{E}_{(x,y) \in q | y\in Y} \max \bar y(x)  \\> \mathbb{E}_{(x,y) \in q | y\in \overline{Y_t}} \max \bar y(x)
\end{split}
\label{eq:prob_hypothesis}
\end{equation}
In other words, the maximal value of the pseudo probability can be used as a measure for identifying the target samples that have labels in $Y$. We, therefore, derive the following scoring mechanism for target samples:
\begin{equation}
    w(x) = d(x) + \max\bar{y}(x)
    \label{eq:weight}
\end{equation}
Let us notice that as $d(x)\in [0,1]$ (higher values are associated with source samples) and $\max\bar{y}(x) \in [0,1]$ it holds that $w(x) \in [0,2]$.

In~\cite{UDA_2019_CVPR}, the authors propose to use a different scoring scheme and use it as a weight for training the second domain classifier they use (we do not employ this component). Their scoring scheme employs the following scores to target domain samples
\begin{equation}
    w_t(x) = d(x)-\frac{H(\overline{y}(x))}{\log |Y_s|}
    \label{eq:uan_weight}
\end{equation}
where $H(\overline{y}(x))$ is the entropy of vector $\overline{y}(x)$. In their work, source domain samples are also scored, by the score $w_s(x)=-w_t(x)$, while we only select target samples as detailed below. Nevertheless, despite using scoring for completely different losses and to different sets of samples, we explore empirically the replacement of our scoring mechanism $w(x)$ with their $w_t(x)$ and demonstrate that our scheme is superior by a sizable margin.

\subsection{Pseudo-labels}
In order to utilize the unlabeled data as much as possible, we opt to use pseudo-labels. As explained in~\cite{choi2019pseudolabeling}, pseudo-labels can be an extremely simple yet effective tool when training a network in a semi-supervised scenario. The difficulty with pseudo-labels in the UDA scenario is the high risk of negative transfer, i.e., decreasing the classifier's performance due to the incorporation of false supervision. In the universal scenario, the target label set is unknown and, therefore, assuming that the network's classification is correct is even more likely to be detrimental than in the conventional domain adaptation case.

In order to deal with the risk of negative transfer, our approach is to use pseudo-labels only on high confidence samples that are likely to be in the shared label set $Y$. As a confidence measure, we employ the sample's transfer score, $w(x)$, and only use pseudo-labels for samples where $w(x)$ is above a certain threshold. We use a dynamic threshold, $w_\alpha(t)$, that changes during the training process according to the following:
\begin{equation}
    w_\alpha(t) = 1.5 - \frac{t}{T}\cdot(1.5 - w_0)
    \label{eq:w_alpha}
\end{equation}
where $t$ is the current training step and $T$ is the total number of training steps. A dynamic threshold is used in order to avoid negative transfer; at first the threshold is set at a high value and as the training advances the network better classifies samples with the threshold $w_0$ and thus it is reasonable to lower the threshold further.

Our pseudo-label classification loss is the following:
\begin{equation}
\begin{split}
    L_C =& \mathbb{E}_{(x,y)\sim p}[L_{CE}(y,\bar{y}(x))] +\\ & \gamma\cdot\mathbb{E}_{(x,y)\sim q}[\mathds{1}_{w(x) > w_\alpha(t)}\cdot L_{CE}(\argmax \bar y(x),\bar y(x))]
    \end{split}
\end{equation}
where $L_{CE}$ is the standard cross-entropy loss, and $\gamma$ is a trade-off parameter.

\subsection{The batch diversity loss}
The method also employs a regularization term aimed at enforcing the samples to be distributed among the different labels in a uniform manner. This regularization helps to better utilize all of the clusters formed in the encoding space, which is likely to be beneficial when transferring to a different  domain.

Given a batch of samples $B = (x_1, x_2, \dots, x_n)$,
for each sample $x_i$ the pseudo probability of it being in class $j$ is denoted by ${\bar{y}(x_i)}_j$.
We define the following regularization term:
\begin{equation}
    \sum_{j=1}^{Y_s} \Big(\frac{1}{|B|}\sum_{i=1}^{|B|}{\bar{y}(x_i)}_j\Big)^2
\end{equation}
This term has a maximal value of 1, which is reached when all of the samples from the batch are mapped to a single class. Its minimal value is $\frac{1}{Y_s}$ and that is obtained if and only if for each sample $\frac{1}{|B|}\sum_{i=1}^{|B|}\bar{y_i}(x_i)_j = \frac{1}{Y_s}$. Adding this penalty to the network's objective encourages a solution that is more uniformly distributed across the different classes.

This loss term is used on samples from both domains. However, in order to avoid negative transfer of samples that are from the target domain specific labels $\overline{Y_t}$, we apply this loss only to samples from the target domain whose sample transfer score $w(x)$ is above a threshold $w_\beta$. Let us denote by $B_s$ the source samples in the batch and $B_t$ the target samples. Let us also denote by $\overline{B_t}$ the target samples for which $w(x) > w_\beta$.  Our loss term becomes the following:
\begin{equation}
\begin{split}
    L_{BD} = \sum_{j=1}^{Y_s} \Big(\frac{1}{|B_s| + |\overline{B_t}|} \cdot 
    \Big(\sum_{i=1,\;x\in p^x}^{|B_s|} \bar{y}(x_i)_j + \\
    \sum_{i=1,\;x\in q^x}^{|B_t|} \mathds{1}_{w(x_i) > w_\beta} \cdot \bar{y}(x_i)_j \Big)\Big)^2
\end{split}
\end{equation}

\subsection{Domain adversarial loss}

 In addition to the losses described above, we also use the conventional adversarial domain loss first introduced in \cite{ganin2016domain}. The domain classifier's network $D$ is trained with a binary cross-entropy loss. 
 \begin{equation}
    L_{D} = \mathbb{E}_{(x,y)\sim p}L_{CE}(1,d(x)) + \mathbb{E}_{(x,y)\sim q}L_{CE}(0,d(x))
 \end{equation}
 A gradient reversal layer is used when backpropagating to network $F$. 

\subsection{The compound loss}
To summarize, the final loss used is the following:
\begin{equation}
    L = L_{C} + L_{BD} - L_{D} 
\end{equation}
Note that the components are unweighted. However, $L_C$ contains the parameter $\gamma$.

\section{Experiments}
We compare our method with state of the art methods from different domain adaptation settings. We also perform a comparison between our scoring mechanism and that proposed in~\cite{UDA_2019_CVPR} and perform an ablation study to show the necessity of our losses.

\noindent{\bf Datasets\quad}
Following~\cite{UDA_2019_CVPR}, we use four datasets. \textbf{Office-Home}~\cite{venkateswara2017deep} is a dataset made up of 65 different classes from four domains: Artistic (Ar), Clipart (Cl), Product (Pr) and Real-world images (RW). Keeping in line with \cite{UDA_2019_CVPR} we test each combination of source and target domain by setting the first 10 classes in alphabetical order as the shared label set $Y$, the next five as the source private, $\overline{Y_s}$, and the rest of the classes (50 classes) are the private target, $\overline{Y_t}$. \textbf{Office31}~\cite{Saenko:2010:AVC:1888089.1888106} consists of three domains, each with 31 classes. The domains are Amazon (\textbf{A}), DSLR (\textbf{D}) and Webcam (\textbf{W}). The 10 shared classes between this dataset and Caltech-256~\cite{caltech256} are used as the shared label set. Aside from these classes, we set the first 10 classes in alphabetical order as $\overline{Y_s}$ and the last 11 classes as $\overline{Y_t}$. \textbf{VisDA2017}~\cite{peng2017visda} is a dataset with a single source and target domain testing the ability to perform transfer learning from synthetic images to natural images. The dataset has 12 classes identical in each domain; we use the first six as the shared label set, the next six as the private source label set and the last three as private target label set. \textbf{ImageNet-Caltech} employs Imagenet-1K~\cite{imagenet_cvpr09} with 1000 different classes and Caltech-256~\cite{caltech256} with 256 classes. The shared label set is comprised of the 84 shared classes between the two datasets, while the source and target private label sets are comprised of all other classes in each dataset.

\noindent{\bf Evaluation protocol\quad}
The protocol of the Open-Set challenge in VisDA2018 is employed. After the training stage, the model is tested only on samples from the target domain. The network must classify the test data into $|Y| + 1$ 
different classes, where the last label $\tau$ contains all labels from the target domain's private label set. As detailed above, our network tries to classify using the labels from the source domain and only classifies into the ``unknown'' class if the sample's transfer score is lower than a predetermined threshold.

\noindent{\bf Implementation details\quad}
The architecture of $F$, $C$, and $D$ follows that of UAN~\cite{UDA_2019_CVPR} in order to provide a direct comparison with this previous work. 
The method is implemented in Pytorch using a ResNet-50 model~\cite{resnet}, pretrained on ImageNet~\cite{imagenet_cvpr09}, as the backbone feature extractor $F$. The label classifier network, $C$, is a fully connected network with a single layer used to classify the features $F(x)$. The domain classifier network, $D$, is comprised of three fully connected layers with ReLU between the first two. 

Our method enjoys a very limited number of hyperparameters. Early on during the development process, we fixed the following hyperparameters across all datasets: $\gamma=0.6$, $w_\beta = 0.8$ and $w_0 = 1.0$. Below we provide some parameter sensitivity experiments to demonstrate the robustness of the method to its parameters.

\subsection{Classification results}
\begin{table*}[t]
\centering
\begin{tabular}{@{}l@{}c@{~}c@{~}c@{~}c@{~}c@{~}c@{~}c@{~}c@{~}c@{~}c@{~}c@{~}c@{~}c@{}}
\hline
\multirow{2}{*}{Method} & \multicolumn{13}{c}{\textbf{Office-Home}} \\ \cline{2-14} 
 & Ar$\shortrightarrow$Cl & Ar$\shortrightarrow$Pr & Ar$\shortrightarrow$Rw & Cl$\shortrightarrow$Ar & {Cl$\shortrightarrow$Pr} & {Cl$\shortrightarrow$Rw} & {Pr$\shortrightarrow$Ar} & {Pr$\shortrightarrow$Cl} & {Pr$\shortrightarrow$Rw} & {Rw$\shortrightarrow$Ar} & {Rw$\shortrightarrow$Cl} & {Rw$\shortrightarrow$Pr} & {Avg} \\ \hline
ResNet~\cite{resnet} & 59.37 & 76.58 & 87.48 & 68.86 & 71.11 & 81.66 & 73.72 & 56.30 & 86.07 & 78.68 & 59.22 & 78.59 & 73.22 \\
DANN~\cite{ganin2016domain} & 56.17 & 81.72 & 85.87 & 68.67 & 73.38 & 83.76 & 69.92 & 56.84 & 85.80 & 79.41 & 57.26 & 78.26 & 73.17 \\
RTN~\cite{DBLP:journals/corr/Long0J16} & 50.46 & 77.80 & 86.90 & 65.12 & 73.40 & 85.07 & 67.86 & 45.23 & 85.50 & 79.20 & 55.55 & 78.79 & 70.91 \\
IWAN~\cite{zhang2018importance} & 52.55 & 81.40 & 86.51 & 70.58 & 70.99 & 85.29 & 74.88 & 57.33 & 85.07 & 77.48 & 59.65 & 79.91 & 73.39 \\
PADA~\cite{cao2018partial} & 39.59 & 69.37 & 76.26 & 62.57 & 67.39 & 77.47 & 48.39 & 35.79 & 79.60 & 75.94 & 44.50 & 78.10 & 62.91 \\
ATI~\cite{busto2019open} & 52.90 & 80.37 & 85.91 & 71.08 & 72.41 & 84.39 & 74.28 & 57.84 & 85.61 & 76.06 & 60.17 & 78.42 & 73.29 \\
OSBP~\cite{saito2018open} & 47.75 & 60.90 & 76.78 & 59.23 & 61.58 & 74.33 & 61.67 & 44.50 & 79.31 & 70.59 & 54.95 & 75.18 & 63.90 \\
UAN~\cite{UDA_2019_CVPR} & 63.00 & 82.83 & 87.85 & 76.88 & 78.70 & 85.36 & 78.22 & \textbf{58.59} & 86.80 & \textbf{83.37} & \textbf{63.17} & 79.43 & 77.02 \\ \hline
Ours & \textbf{63.59} & \textbf{85.02} & \textbf{91.42} & \textbf{77.01} & \textbf{84.09} & \textbf{88.29} & \textbf{79.50} & 56.49 & \textbf{89.85} & 77.52 & 61.00 & \textbf{85.69} & \textbf{78.29} \\ \hline
\end{tabular}
\smallskip\caption{ Average class accuracy (\%) on the Office-Home ($\xi = 0.15$). The results for all methods besides our are taken from UAN\cite{UDA_2019_CVPR}}
\label{table:res1}
\end{table*}

\begin{table*}[t]
\centering
\begin{tabular}{@{}l@{}c@{~}c@{~}c@{~}c@{~}c@{~}c@{~}c@{~}c@{~}c@{~}c@{~}c@{~}c@{~}c@{}}
\hline
\multirow{2}{*}{Method} & \multicolumn{7}{c|}{\textbf{Office31}} & \multicolumn{3}{c|}{\textbf{ImageNet-Caltech}} & \multicolumn{1}{l}{\multirow{2}{*}{\textbf{VisDA2017}}} \\ \cline{2-11}
 & A $\rightarrow$ W & D $\rightarrow$ W & W $\rightarrow$ D & A $\rightarrow$ D & \multicolumn{1}{l}{D $\rightarrow$ A} & \multicolumn{1}{l}{W $\rightarrow$ A} & \multicolumn{1}{l|}{Avg} & \multicolumn{1}{l}{I $\rightarrow$ C} & \multicolumn{1}{l}{C $\rightarrow$ I} & \multicolumn{1}{l|}{Avg} & \multicolumn{1}{l}{} \\ \hline
ResNet~\cite{resnet} & 75.94 & 89.60 & 90.91 & 80.45 & 78.83 & 81.42 & \multicolumn{1}{c|}{82.86} & 70.28 & 65.14 & \multicolumn{1}{c|}{67.71} & 52.80 \\
DANN~\cite{ganin2016domain} & 80.65 & 80.94 & 88.07 & 82.67 & 74.82 & 83.54 & \multicolumn{1}{c|}{81.78} & 71.37 & 66.54 & \multicolumn{1}{c|}{68.96} & 52.94 \\
RTN~\cite{DBLP:journals/corr/Long0J16} & 85.70 & 87.80 & 88.91 & 82.69 & 74.64 & 83.26 & \multicolumn{1}{c|}{84.18} & 71.94 & 66.15 & \multicolumn{1}{c|}{69.05} & 53.92 \\
IWAN~\cite{zhang2018importance} & 85.25 & 90.09 & 90.00 & 84.27 & 84.22 & 86.25 & \multicolumn{1}{c|}{86.68} & 72.19 & 66.48 & \multicolumn{1}{c|}{69.34} & 58.72 \\
PADA~\cite{cao2018partial} & 85.37 & 79.26 & 90.91 & 81.68 & 55.32 & 82.61 & \multicolumn{1}{c|}{79.19} & 65.47 & 58.73 & \multicolumn{1}{c|}{62.10} & 44.98 \\
ATI~\cite{busto2019open} & 79.38 & 92.60 & 90.08 & 84.40 & 78.85 & 81.57 & \multicolumn{1}{c|}{84.48} & 71.59 & 67.36 & \multicolumn{1}{c|}{69.48} & 54.81 \\
OSBP~\cite{saito2018open} & 66.13 & 73.57 & 85.62 & 72.92 & 47.35 & 60.48 & \multicolumn{1}{c|}{67.68} & 62.08 & 55.48 & \multicolumn{1}{c|}{58.78} & 30.26 \\
UAN~\cite{UDA_2019_CVPR} & 85.62 & 94.77 & \textbf{97.99} & 86.50 & 85.45 & 85.12 & \multicolumn{1}{c|}{89.24} & 75.28 & 70.17 & \multicolumn{1}{c|}{72.73} & 60.83 \\ \hline
Ours & \textbf{90.25} & \textbf{95.25} & 96.96 & \textbf{88.84} & \textbf{90.19} & \textbf{89.30} & \multicolumn{1}{c|}{\textbf{91.80}} & \textbf{76.13} & \textbf{74.67} & \multicolumn{1}{c|}{\textbf{75.40}} & \textbf{64.31} \\ \hline
\end{tabular}
\smallskip\caption{Average class accuracy for Office31($\xi=0.32$), ImageNet-Caltech ($\xi=0.07$) and VisDA2017($\xi = 0.50)$}
\label{table:res2}
\end{table*}

We compare our approach with prior methods in the UDA setting. Tab.~\ref{table:res1},~\ref{table:res2} present the results on the acceptable benchmarks of the field. The success rate for methods other than ours is taken from~\cite{UDA_2019_CVPR}. As can be observed, our approach achieves state of the art results on the majority of the domain adaptation tasks across the different datasets. Our method is able to improve the average classification accuracy of each dataset by 1-5 percents.

\subsection{Scoring scheme analysis}

\begin{table*}[t]
\centering
\begin{tabular}{lccccccc}
\hline
 & A $\rightarrow$ W & D $\rightarrow$ W & W $\rightarrow$ D & A $\rightarrow$ D & \multicolumn{1}{l}{D $\rightarrow$ A} & \multicolumn{1}{l}{W $\rightarrow$ A} & \multicolumn{1}{l}{Avg} \\ \hline
UAN\cite{UDA_2019_CVPR} & 85.62 & 94.77 & \textbf{97.99} & 86.50 & 85.45 & 85.12 & 89.24\\ \hline 
Ours with $w_t(x)$ & 86.23 & 93.26 & 91.79 & 84.31 & 86.09 & 85.41 & 87.84 \\
Ours with ${w_h}(x)$ & 85.26 & 93.81 & 95.16 & 82.84 & 85.31 & 83.51 & 87.65 \\
Ours, $w(x)$ w/o $d(x)$ & 89.95 & \textbf{95.70} & 96.46 & 88.05 & 89.92 & 89.05 & 91.52  \\
Ours, $w(x)$ w/o $\max{\bar{y}(x)}$ & 81.9 & 89.11 & 87.85 & 83.68 & 81.73 & 82.55 & 82.55 \\
Ours with $w(x)$ &  \textbf{90.25} & 95.25 & 96.96 & \textbf{88.84} & \textbf{90.19} & \textbf{89.30} & \textbf{91.80} \\\hline
\end{tabular}
\smallskip\caption{Comparison on Office31 between UAN\cite{UDA_2019_CVPR} and our approach when using either $w(x)$ (with ablation on the score's components) or UAN's $w_t(x)$ as the scoring scheme, as well as other variants.}
\label{table:weight_compare}
\end{table*}

\begin{figure*}
  \centering
    \begin{tabular}{ccc}
  \includegraphics[width=0.25\textwidth]{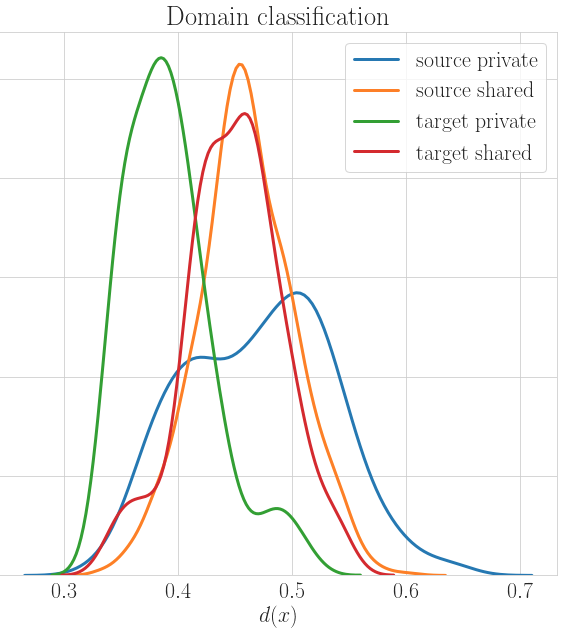} & \includegraphics[width=0.25\textwidth]{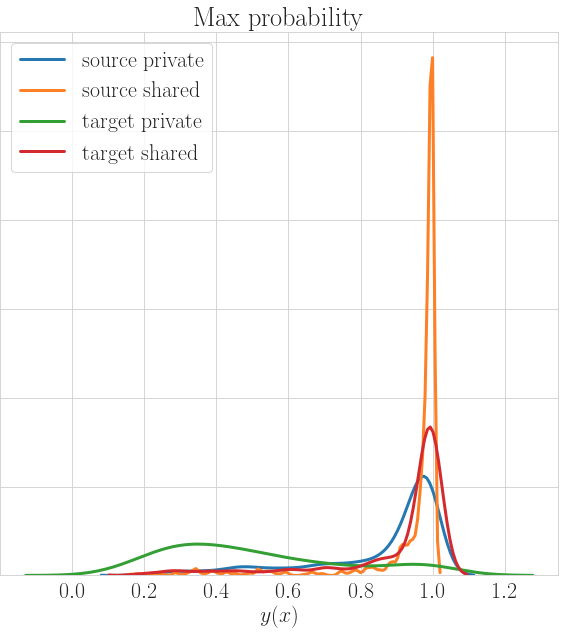} &
  \includegraphics[width=0.25\textwidth]{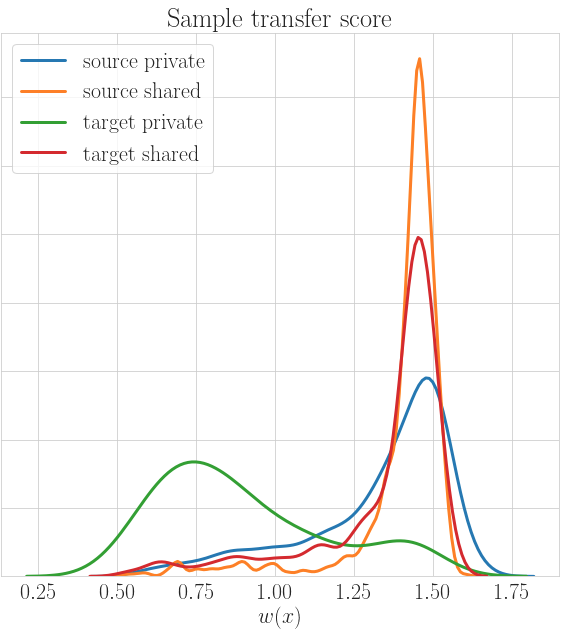} \\
  (a) & (b) & (c)
  \end{tabular}
  \smallskip\caption{Distributions of the different components of the scoring scheme on the four following sample groups: source samples in $Y$ (orange), source sample in $\overline{Y_s}$ (blue), target samples in $Y$ (red) and target samples in $\overline{Y_t}$ (green). (a) Distribution of the domain classifier's output $d(x)$. (b) The label classifier's maximum probability, $\max\bar{y}(x)$. (c) The score $w(x)$, which combines both.}
\label{fig:weight_componenets}
\end{figure*}

In Fig.~\ref{fig:weight_componenets} we present the estimated probability density function for the different components of $w(x)$ on the Office31 dataset for the domain shift \textbf{A}$\rightarrow$\textbf{D}. $d(x)$, shown in Fig.~\ref{fig:weight_componenets}(a), displays the following expected behavior: $\mathbb{E}_{(x,y) \in p | y \in \overline{Y_s}} d(x) > \mathbb{E}_{(x,y) \in p | y \in Y} d(x) \approx \mathbb{E}_{(x,y) \in q | y\in Y} d(x) > \mathbb{E}_{(x,y) \in q | y\in \overline{Y_t}} d(x)$. In Fig.~\ref{fig:weight_componenets}(b) we analyze the max probability of the classifier, $\max\bar{y}(x)$, validating the hypothesis in Eq.~\ref{eq:prob_hypothesis} and justifying using this component as part of our score scheme. 
Finally, in Fig.~\ref{fig:weight_componenets}(c), we present the full sample transfer score $w(x)$.  The results show that target samples with higher scores $w(x)$ are typically from the shared label set. This justifies the use of our scoring scheme to distinguish between samples that we can predict correctly and those that should be labeled $\tau$.

\paragraph{Comparing scoring schemes}We next compare our proposed scoring scheme $w(x)$, as shown in Eq.~\ref{eq:weight}, to the scoring scheme proposed in \cite{UDA_2019_CVPR}, $w_t(x)$ given by Eq.~\ref{eq:uan_weight}. In order to compare the two scoring schemes, we use the score $w_t(x)$ proposed on the target samples instead of $w(x)$. The method that uses $w_t(x)$ was tuned to optimize its performance. 
In addition to the scoring scheme $w_t(x)$, we also compare to an entropy based one, since entropy has been shown to be a good criterion in domain adaptation~\cite{Grandvalet2004entropy,long2016unsupervised}. Based on the assumptions that target samples from the shared label set are similar to the source samples and will thus have a lower entropy, we define the following scoring scheme:
\begin{equation}
    {w_h}(x) = 1 - \frac{H(\overline{y}(x))}{\log |Y_s|}
    \label{eq:entropy_score}
\end{equation}
The comparison is done on Office31 dataset and the results appear in Tab.~\ref{table:weight_compare}. As can be seen, our scoring scheme produces superior results across the entire dataset. We thus conclude that our scoring mechanism outperforms the one proposed in \cite{UDA_2019_CVPR} and ${w_h}(x)$ when used in the context of our method.

Tab.~\ref{table:weight_compare} also presents an ablation study on the components of the scoring mechanism. ``$w(x)$ w/o $d(x)$'' refers to the score function when removing the domain factor $d(x)$ from Eq.~\ref{eq:weight} and ``$w(x)$ w/o $\max{\bar{y}(x)}$'' to the score function when removing the classification component. The results show that both components are necessary for achieving our final result. However, the classification component is more crucial to the success of our scoring mechanism, and by itself already outperforms the state of the art.

\subsection{Parameter sensitivity}

\begin{figure}
  \centering
    \begin{tabular}{c}
  \includegraphics[width=0.995\linewidth]{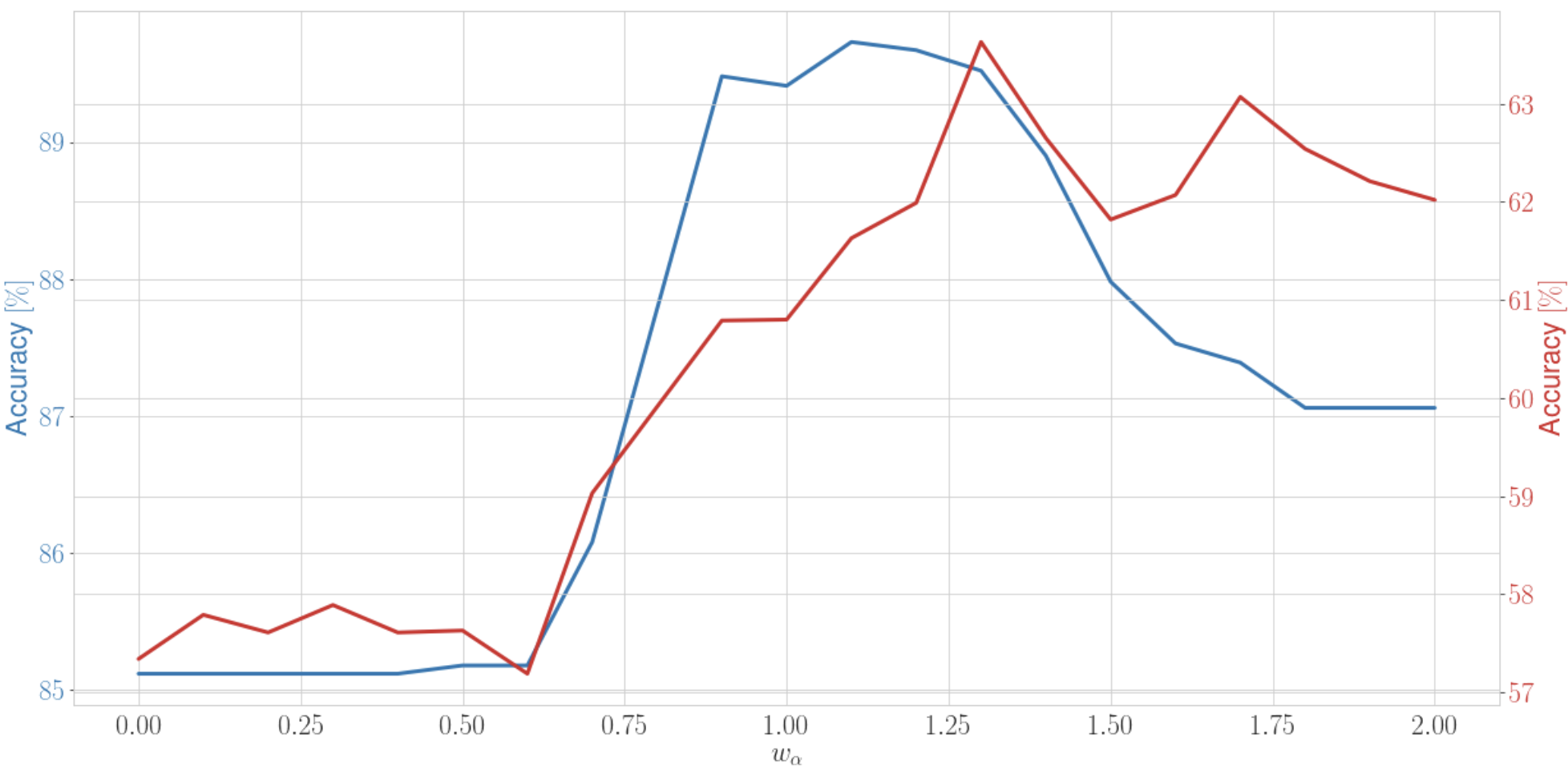}
  \end{tabular}
  \smallskip\caption{(blue) Accuracy on Office31 \textbf{A}$\rightarrow$\textbf{D} ($w_\beta = 1.0$, $w_0 = 1.0$, $\gamma=0.5$) and (red) on OfficeHome \textbf{Ar}$\rightarrow$\textbf{Cl} ($w_\beta = 0.8$, $w_0 = 1.0$, $\gamma=0.6$) as a function of the threshold $w_\alpha$. Note that there are two $y$ axes due to a different level of baseline performance.}
\label{fig:pl_thresh}
\end{figure}

\noindent{\bf Pseudo-label threshold analysis\quad}
\begin{figure}
  \centering
    \begin{tabular}{@{}c@{~}c@{}} 
    \includegraphics[width=0.495\linewidth]{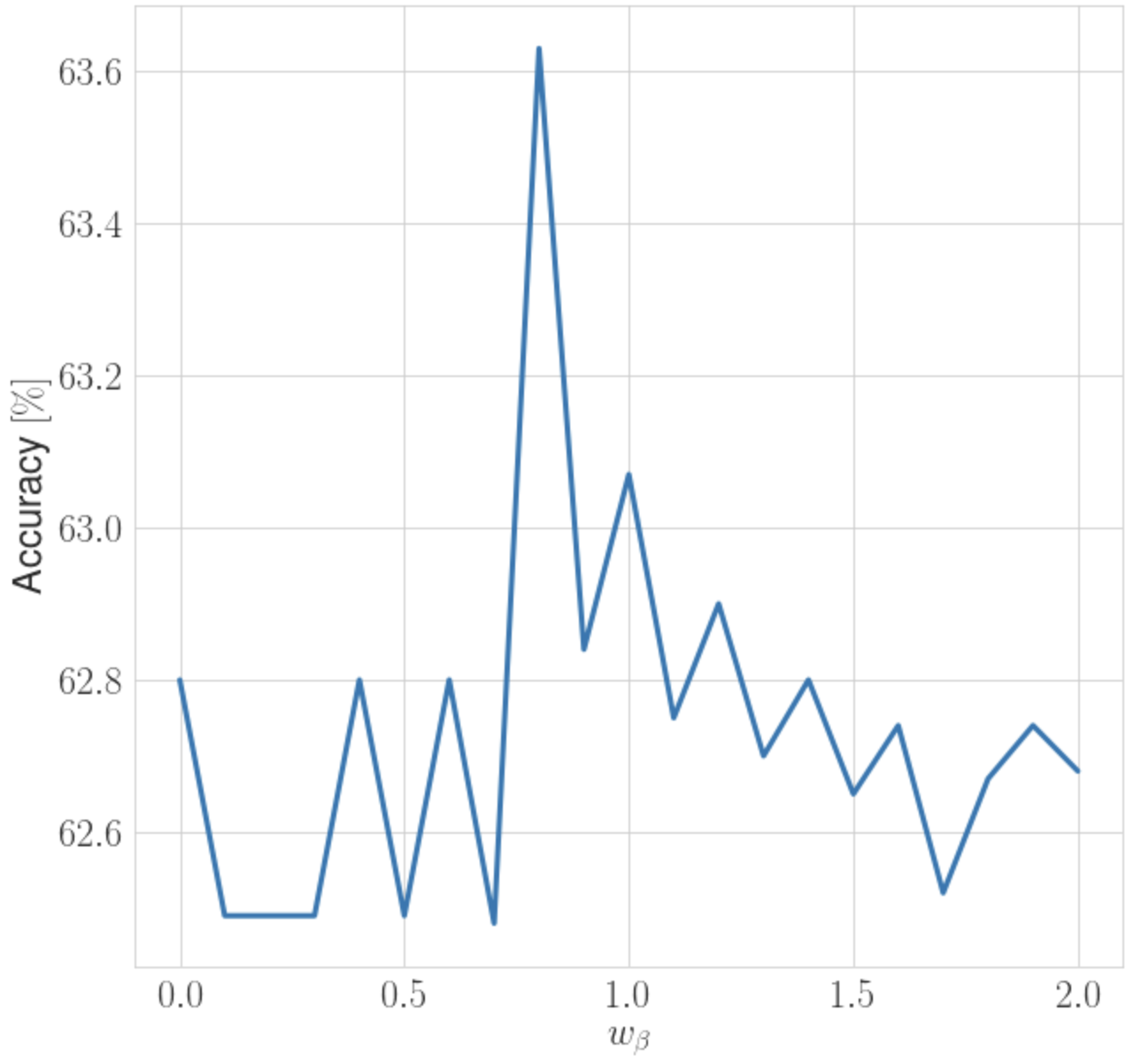}&
  \includegraphics[width=0.495\linewidth]{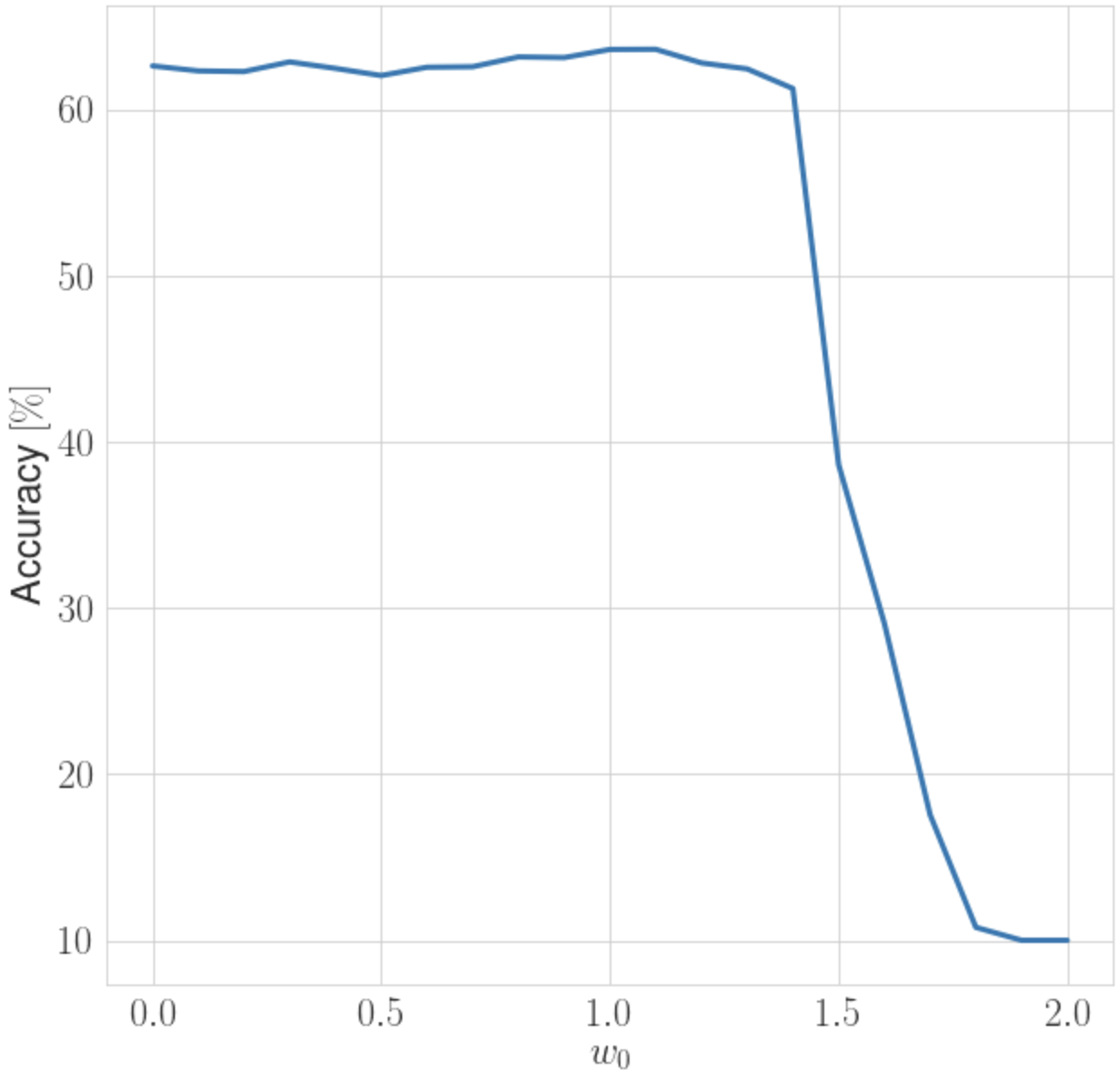}\\
  (a) & (b)\\
  \end{tabular}
  \smallskip\caption{(a) Accuracy w.r.t. threshold $w_\beta$ ($w_0 = 1.0$, $w_\alpha=1.3$, $\gamma=0.6$). (b) Accuracy w.r.t $w_0$ ($w_\beta = 0.8$, $\gamma=0.6$, dynamic $w_\alpha$). Both results are on OfficeHome \textbf{Art} to \textbf{Clipart}}
\label{fig:thresh}
\end{figure}
We study the sensitivity of our method to the threshold $w_\alpha$, which is used to determine whether or not the pseudo-label of a target sample should be taken into consideration when calculating its loss. While our method employs a dynamic threshold, in order to obtain a clearer image, we perform the experiment when the threshold is fixed. We compare the average accuracy on the OfficeHome dataset with the domain shift \textbf{Ar}$\rightarrow$\textbf{Cl} and on Office31 with the domain shift \textbf{A}$\rightarrow$\textbf{D}. The tests are conducted by fixing all other hyperparameters to the default values and only changing the value of $w_\alpha$. 

The results are presented in Fig.~\ref{fig:pl_thresh}. By taking the lowest threshold possible, $w_\alpha = 0$, we allow the use of pseudo-labels on every sample seen during the training stage. As can be seen from the performance graph, this yields a much lower result than higher thresholds, probably due to negative transfer. This result is also evident when looking at the results of Tab.~\ref{table:pl_ablation_office31},~\ref{table:dynamic_thresh} when setting the threshold $w_\alpha = 0$.

The second edge case is $w_\alpha = 2$, which is the maximal value that the $w(x)$ can have. With this threshold, no score will ever satisfy $w(x) > w_\alpha$ and thus it is equivalent to not using pseudo-labels at all. From Fig.~\ref{fig:pl_thresh} one can observe that $w_\alpha = 2$ does not yield the best results, meaning that the use of pseudo-labels does, in fact, help train the network. Tab.~\ref{table:dynamic_thresh} shows the results on the OfficeHome dataset in the case where pseudo-labels are applied as suggested in our approach and when they are not applied at all. These results show that the use of pseudo-labels during training does improve the accuracy during the deployment stage across the entire dataset. These results are also affirmed by Tab.~\ref{table:pl_ablation_office31} for the Office31 tasks.

In addition, one can observe that setting a very low threshold also leads to results much lower than the best possible performance. This comes as no surprise, since using pseudo-labels on all the target samples leads to negative transfer. 

\noindent{\bf Dynamic threshold\quad}
\begin{table*}[t]
\centering
\begin{tabular}{lccccccc}
\hline
 & A $\rightarrow$ W & D $\rightarrow$ W & W $\rightarrow$ D & A $\rightarrow$ D & \multicolumn{1}{l}{D $\rightarrow$ A} & \multicolumn{1}{l}{W $\rightarrow$ A} & \multicolumn{1}{l}{Avg} \\ \hline
Ours w/o pseudo-labels & 85.38 & 94.34 & 96.67 & 87.58 & 85.00 & 85.32 & 89.05 \\
Ours, $w_\alpha = 0$ & 87.92 & 90.62 & 87.44 & 85.19 & 84.00 & 84.65 & 86.64 \\
Ours, static $w_\alpha = 1.2$ & 89.75 & \textbf{95.47} & 95.95 & 88.78 & 89.53 & 88.88 & 91.39 \\
Ours & \textbf{90.25} & 95.25 & \textbf{96.96} & \textbf{88.84} & \textbf{90.19} & \textbf{89.30} & \textbf{91.80} \\ \hline
\end{tabular}
\smallskip\caption{Comparison between different thresholds for the use of pseudo-labels on Office31.}
\label{table:pl_ablation_office31}
\medskip
\centering
\begin{tabular}{@{}l@{}c@{~}c@{~}c@{~}c@{~}c@{~}c@{~}c@{~}c@{~}c@{~}c@{~}c@{~}c@{~}c@{}}
\hline
 & Ar$\shortrightarrow$Cl & Ar$\shortrightarrow$Pr & Ar$\shortrightarrow$Rw & Cl$\shortrightarrow$Ar & {Cl$\shortrightarrow$Pr} & {Cl$\shortrightarrow$Rw} & {Pr$\shortrightarrow$Ar} & {Pr$\shortrightarrow$Cl} & {Pr$\shortrightarrow$Rw} & {Rw$\shortrightarrow$Ar} & {Rw$\shortrightarrow$Cl} & {Rw$\shortrightarrow$Pr} & {Avg} \\ \hline
Ours w/o pseudo-labels & 63.0 & 83.97 & 90.53 & 72.40 & 75.56 & 84.25 & 70.83 & 54.21 & 89.01 & 74.96 & 56.76 & 84.20 & 74.97 \\
Ours, $w_\alpha = 0$ & 57.34 & 81.50 & 88.51 & 72.58 & 83.06 & 86.32 & 75.73 & 58.46  & 86.94 & 74.84 & 58.33 & 82.29 & 75.49 \\
Ours, static $w_\alpha = 1.2$ & 62.93 & 82.86 & 90.60 & 74.52 & \textbf{84.27} & \textbf{88.84} & \textbf{81.66} & 55.10 & 88.92 & \textbf{78.56} & \textbf{61.44} & 84.25 & 77.83 \\
Ours & \textbf{63.59} & \textbf{85.02} & \textbf{91.42} & \textbf{77.01} & 84.09 & 88.29 & 79.50 & \textbf{56.49} & \textbf{89.85} & 77.52 & 61.00 & \textbf{85.69} & \textbf{78.29} \\ \hline
\end{tabular}
\smallskip\caption{Comparison between different thresholds for the use of pseudo-labels on Office-Home.
\label{table:dynamic_thresh}}
\medskip
\centering
\begin{tabular}{lccccccc}
\hline
 & A $\rightarrow$ W & D $\rightarrow$ W & W $\rightarrow$ D & A $\rightarrow$ D & \multicolumn{1}{l}{D $\rightarrow$ A} & \multicolumn{1}{l}{W $\rightarrow$ A} & \multicolumn{1}{l}{Avg} \\ \hline
No diversity loss & 87.97 & \textbf{95.77} & \textbf{97.11} & 88.20 & 89.16 & 88.53 & 91.12 \\
Diversity loss on target samples& 85.68 & 95.22 & 96.58 & 87.81 & 89.29 & 88.25 & 90.47 \\
Diversity loss on all samples & \textbf{90.25} & 95.25 & 96.96 & \textbf{88.84} & \textbf{90.19} & \textbf{89.30} & \textbf{91.80} \\ \hline
\end{tabular}
\smallskip\caption{Results on Office31 when using the diversity loss in different manners. 
\label{table:diversity_loss_office31}}
\medskip
\centering
\begin{tabular}{@{}l@{}c@{~}c@{~}c@{~}c@{~}c@{~}c@{~}c@{~}c@{~}c@{~}c@{~}c@{~}c@{~}c@{}}
\hline
 & Ar$\shortrightarrow$Cl & Ar$\shortrightarrow$Pr & Ar$\shortrightarrow$Rw & Cl$\shortrightarrow$Ar & {Cl$\shortrightarrow$Pr} & {Cl$\shortrightarrow$Rw} & {Pr$\shortrightarrow$Ar} & {Pr$\shortrightarrow$Cl} & {Pr$\shortrightarrow$Rw} & {Rw$\shortrightarrow$Ar} & {Rw$\shortrightarrow$Cl} & {Rw$\shortrightarrow$Pr} & {Avg} \\ \hline
No diversity loss & \textbf{63.81} & 84.65 & 91.42 & \textbf{77.65} & 84.15 & 87.81 & \textbf{79.92} & \textbf{56.71} & 88.11 & 76.6 & 60.24 & 85.43 & 78.04  \\
Target only & 63.04 & \textbf{85.03} & \textbf{91.49} & 77.07 & \textbf{84.47} & 87.78 & 79.36 & 56.42 & 88.80 & 77.32 & 60.21 & 85.61 & 78.05  \\
Target + Source & 63.59 & 85.02 & 91.42 & 77.01 & 84.09 & \textbf{88.29} & 79.50 & 56.49 & \textbf{89.85} & \textbf{77.52} & \textbf{61.00} & \textbf{85.69} & \textbf{78.29} \\ \hline
\end{tabular}
\smallskip\caption{Results on OfficeHome when using the diversity loss in different manners.%
\label{table:diversity_loss}}
\end{table*}
We next analyze the advantage of employing a dynamic threshold for the use of pseudo-labels. The analysis is done on the Office31 dataset by fixing a set threshold,  $w_\alpha = 1.20$ (which was found to provide the optimal value). 

The results are reported in Tab.~\ref{table:dynamic_thresh}. As can be seen, the use of a dynamic threshold does seem to give better results overall. This is probably due to the fact that we are able to use more samples whose transfer score $w(x)$ is above $w_0$ and below the set $w_\alpha$ at later parts of the training. 

Another insight to the advantage of using a dynamic threshold can be seen by looking at Fig.~\ref{fig:pl_thresh}. From the figure, one can observe that different domains and datasets give better results when using different thresholds for the pseudo-labels. In particular, for the domain shift \textbf{Ar}$\rightarrow$\textbf{Cl} in OfficeHome, the best result is achieved with $w_\alpha=1.3$ and for Office31 \textbf{A}$\rightarrow$\textbf{D}, the best result is achieved at $w_\alpha = 1.1$. The dynamic threshold enables us to take advantage of different thresholds during training, yielding improved results over a number of datasets.

\noindent{\bf Diversity loss threshold analysis\quad}
We next analyze the threshold $w_\beta$ used to determine which of the target samples are used when calculating the batch diversity loss. We compare the average accuracy when only changing the threshold $w_\beta$ while all other hyper-parameters are fixed at the default value (including the threshold $w_\alpha$).  The results can be found in Fig.~\ref{fig:thresh}(a). The accuracy varies by around 1\% but it is clear that the use of pseudo-labels under a relatively stable threshold value does improve the final result. Note that the default threshold is not the one that produces the best possible outcome.

\noindent{\bf Decision threshold analysis\quad}
Another component of the network we analyze is the decision threshold $w_0$, which is used to decide whether the model would label a sample as $\tau$ or use the predicted label. The analysis is done in a similar manner to the two previous sections. However, here we use a dynamic threshold as described in Eq.~\ref{eq:w_alpha}.

As is evident from the results in Fig.~\ref{fig:thresh}(b), there is little variance in the results for a threshold in a wide range between $0$ and $1.4$. For thresholds higher than $1.4$, we see a sharp fall in the accuracy until finally reaching the lowest possible values at $w_0=2$. This fall in accuracy occurs because for high enough thresholds only a very small number of samples have a transfer score $w(x)$ higher than the threshold and thus most samples are labeled $\tau$. The extreme case, as seen in the graph, is $w_0=2$ where no sample can pass this threshold and all are labeled $\tau$, leading to an accuracy score that is $\xi$ the fraction of samples from novel target classes in this benchmark.

\paragraph{Diversity loss study}
We next explore the contribution of the batch diversity loss and whether the diversity loss only needs to be applied on the target domain, for which we have no label, as a means of semi-supervised learning, or whether using it also on the source domain helps improve the network's outcome. All hyper-parameters used in the analysis are the same default parameters as described above. 

Tab.~\ref{table:diversity_loss_office31}~(\ref{table:diversity_loss}) presents the results on the Office31 (OfficeHome) dataset in three scenarios: (i) when not applying the diversity loss at all, (ii) when using the diversity loss on both domains and (iii) when using the loss only on the target domain. As can be seen, the results for the different settings are very similar. However,  the use of the batch diversity loss on both domains does yield slightly better results. 

\section{Conclusions}

We study unsupervised domain adaptation in the challenging case where there is a partial overlap between the source and target domain classes. Our method adapts through the usage of pseudo-labels and a diversity loss. However, since some of the samples of the target domain cannot be properly labeled by any of the source labels, we propose to score the samples and apply a threshold. 

Our scoring takes into consideration 
the confidence of the label classifier, as well as the confidence of the domain discriminator. The more certain the first network is in its prediction and the less certain the second is that the sample is from the target domain, the more likely the target domain sample is from the shared label set.

The method obtains state of the art results by a sizable margin on the relevant literature benchmarks, despite being simpler than previous work. We also demonstrate that our scoring scheme is superior to the ones previously proposed.

\subsection*{Acknowledgements}
This project has received funding from the European Research Council (ERC) under the European Union’s Horizon 2020 research and innovation programme (grant ERC CoG 725974). 

{ \small
\bibliographystyle{ieee_fullname}
\bibliography{gans}
}

\end{document}